\documentclass[10pt]{article}
\usepackage[margin=1in]{geometry}

\usepackage[colorlinks,citecolor=red]{hyperref}

\usepackage{graphicx}
\DeclareGraphicsExtensions{.pdf}
\usepackage[justification=Centering,singlelinecheck=false,caption=false,font=footnotesize]{subfig}

\usepackage{natbib}
\bibpunct{(}{)}{;}{a}{}{,}

\usepackage[cmex10]{amsmath}
\usepackage{amsfonts}
\usepackage{bm}
\numberwithin{equation}{section}

\usepackage{amsmath}
\usepackage{amsfonts}
\usepackage{amssymb}
\usepackage{color}

\usepackage{graphicx}
\usepackage{url}
\usepackage{algorithm}
\usepackage{algorithmic}
\usepackage{subfig}
\usepackage{array}

\usepackage{multirow}
\usepackage{array}
\renewcommand{\arraystretch}{1.5}
\usepackage{makecell}

\usepackage{hyperref}
\usepackage{cleveref}

\begin{document}

\title{\textbf{Graph Federated Unlearning for Privacy Preservation}}

\author{Ruotong Ma \hspace{1cm} Wentao Yu \hspace{1cm} Qizhou Wang \hspace{1cm} Jie Yang \hspace{1cm} Chen Gong\\
Shanghai Jiao Tong University\\Nanjing University of Science and Technology\\Hong Kong Baptist University\\China\\
Email: ruotong0830@gmail.com, blgpb.ywt@gmail.com, 
csqzwang@comp.hkbu.edu.hk}

\date{29 January 2026}

\maketitle

\begin{abstract}
  \emph{Graph federated learning} (GFL) facilitates \textcolor{black}{decentralized} training on distributed graph data while keeping sensitive user information local, aligning with policies such as GDPR and CCPA that grant users the right to freely join or withdraw from learning systems. 
  However, even decentralized, user information can persist after quitting, potentially propagating to \textcolor{black}{central servers} and then redistributing to malicious \textcolor{black}{clients}. This \textcolor{black}{privacy leakage} during \textcolor{black}{user withdrawal}, despite its importance, has received seldom attention in GFL. To fill the gap, we explore the potential of \emph{machine unlearning} (MU) to thoroughly remove \textcolor{black}{user information}. However, classical MU methods are known to \textcolor{black}{degrade overall performance}, a problem that is exacerbated in GFL due to \textcolor{black}{local message passing} and \textcolor{black}{global model collaboration}. To this end, we make two adjustments to mitigate this challenge for GFL. 
  First, we ensure unlearning updates that minimally affect overall performance, steering them in directions orthogonal to the gradients from learning other data. 
  Second, we introduce virtual clients, maintained by the central server, to preserve graph topology and global embeddings without recovering information of removed entities. 
  We conduct comprehensive experiments under a representative user-withdrawal scenario and propose a novel membership inference framework to rigorously evaluate and validate the reliability of our privacy preservation. The experimental results demonstrate the effectiveness of our approach, which also surpasses the performance of seven state-of-the-art baseline methods.
\end{abstract}

\section{Introduction} \label{introduction}

Graph-structured data are pervasive in modern AI applications, and {\emph{graph neural networks} (GNNs)}~\cite{xu2018powerful,wu2020comprehensive,luan2022revisiting,corso2024graph,yu2026atom} have been extensively studied as a powerful architecture for analyzing such data, leveraging message-passing mechanisms to capture both \textcolor{black}{node attributes} and inherent \textcolor{black}{topological structures}. Further with recent advances in spectral graph theory~\cite{bo2023surveyspectralgraphneural} and attention mechanisms~\cite{ma2025graphattentionbeneficialtheoretical}, GNNs are now capable of  supporting large-scale, real-world applications ranging from high-frequency financial trading~\cite{li2024anomaly} to industrial recommender systems~\cite{ko2022survey}. However, processing such large volumes of data with \textcolor{black}{message propagation}, especially in data-sensitive domains, makes GNNs particularly susceptible to privacy risks, thereby hindering their broader deployments~\cite{liu2021machine}.

In response to increasingly stringent data-sharing and privacy regulations, \emph{graph federated  learning} (GFL)~\cite{zhou2024traffic, wentao2025fediih} has emerged as a promising solution. GFL enables GNN training on clients, further with these local-trained models collaboratively contributing to a central server. Without directly exchanging data across clients, this decentralized mechanism aligns better with mainstream data-sharing regulations~\cite{voigt2017eu}. However, while these privacy-preserving improvements are remarkable, they do not resolve all concerns, namely gradient information and model parameters remain vulnerable to membership inference~\cite{liu2021machine} and inversion attacks~\cite{he2019model}, thereby still posing privacy risks and necessitating further advancements. In this paper, we focus on an important yet seldom-discussed implication of this vulnerability: 
\textit{GFL cannot fully satisfy the ``right to be forgotten'' (RTBF)}~\cite{voigt2017eu}.

Generally speaking, RTBF grants individuals the right to request the deletion of their \textcolor{black}{personal information} from online databases and/or AI systems. It is a cornerstone of modern data protection laws, most notably the GDPR~\cite{voigt2017eu}, and has been adopted by following regulations such as CCPA~\cite{bonta2022california}, LGPD~\cite{de2025proteccao}, and DPDPA~\cite{dave2025comparative}. In the context of GFL, RTBF implies that users can freely join or withdraw from the system, with their associated information being embedded into or removed from GNN parameters accordingly. At first glance, it may seem that GFL systems can easily achieve this requirement by removing the corresponding user clients. However, as we will show in Section~\ref{mainresult}, membership inference~\cite{liu2021machine} reveals that associated information can still be detected from other clients, since local parameters are propagated and embedded within other models. 


To mitigate this privacy leakage issue during user withdrawal, we explore the potential use of \emph{machine unlearning} (MU)~\cite{cao2015towards} for RTBF. Classical MU methods typically assume that a small portion of training data needs to be removed and aim to eliminate its influence, giving rise to \emph{differential privacy} (DP)-based approaches~\cite{bourtoule2021machine,chen2021machine,tarun2023fast,wang2025gru}. 
However, these methods have been shown to incur prohibitive computational costs for large-scale models and to degrade overall performance, motivating researchers to investigate \emph{gradient-ascent} (GA)-based methods~\cite{fan2023salun,huang2024unified,liu2025rethinking,wang2025rethinking,li2025llm} that directly reduce the likelihood of generating predictions for data to be unlearned. While GA-based methods are easier to implement, they still face the critical challenge of harming overall model performance. 
This issue can be even more severe for GFL, as the negative impact of unlearning may \textcolor{black}{propagate and accumulate across local nodes} and further spread through \textcolor{black}{federated model redistribution}, ultimately harming clients that are not intended to be unlearned. Therefore, existing GA-based methods need to be further improved to better satisfy RTBF requirements in the context of GFL.

To this end, we propose a \emph{graph \underline{\textbf{Fed}}erated unlearning method with \underline{\textbf{G}}radient \underline{\textbf{C}}orrection and \underline{\textbf{V}}irtual client (FedGCV)}, an enhanced GA-based MU framework tailored for graph federated unlearning (GFU). To mitigate the negative impact on overall performance, FedGCV introduces two key improvements over previous MU methods. 
First, to counteract the erroneous propagation of unlearning effects to \textcolor{black}{normal clients} that are not intended to be removed, we introduce \textbf{gradient correction} to explicitly preserve overall GNN performance. Specifically, we constrain the unlearning gradients to be orthogonal to that used for learning other clients, which can be coordinated by the central servers and justified as \textcolor{black}{performance-preserving} under a first-order approximation.
Second, we further preserve topological structures by introducing \textbf{virtual clients}, also organized by the central server, to replace the unlearned clients and synthesize graphs to maintain both node attributes and inherent topological structures. The virtual clients preserve graph topologies and global embeddings without leaking information of removed entities, which can be employed to preserve the performance of the global model. Therefore, GFU combines gradient correction and virtual clients, aiming to enhance performance preservation while maintaining reliable RTBF.

We further propose a membership-inference–based evaluation framework in~\cref{mia-eva} to assess the degree of privacy preservation. Moreover, this is the first work to employ MIA in the GFU scenario. Through extensive experiments across a representative RTBF setup for GFU, we demonstrate the reliability of our method in preserving overall model performance meanwhile mitigating privacy leakage.

\section{Preliminaries}
In this section, we briefly review the relevant background. First, we introduce GNNs for graph-structured data, in Section~\ref{sec:pre:gnn}. Then, we discuss the use of federated learning with GNNs, leading to GFL, in Section~\ref{sec:pre:fl}. Finally, we highlight the drawbacks of GFL, particularly its limited flexibility in meeting advanced data-sharing regulations, in Section~\ref{sec:pre:rtbf}.


\subsection{Graph Neural Networks}\label{sec:pre:gnn}

GNNs extend conventional deep neural networks to handle graph-structured data via its unique message-passing mechanism~\cite{bai2022two, 10032180}. In general, GNNs adopt a neighborhood aggregation strategy, in which the representation of a node $v_i$ is iteratively updated by aggregating that of its neighborhoods. Therefore, after $k$ iterations of aggregation, the representation of the node $v_i$ captures the structural information contained within its $k$-hop neighboring nodes. Formally, the learned representation of $v_i$ at the $k$-th layer, denoted by $h_{v_i}^{(k)}$, can be written as
\begin{equation}\label{eq1}
h_{v_i}^{(k)}=\operatorname{CO}^{(k)}\big(h_{v_i}^{(k-1)}, \operatorname{AG}^{(k)}(\{h_{v_j}^{(k-1)}:v_j\in\mathcal{N}(v_i)\})\big).
\end{equation}
Therein, $\mathcal{N}({v_i})$ denotes the set of neighboring nodes of $v_i$. \(\mathrm{AG}^{(k)}\) aggregates messages from $\mathcal{N}({v_i})$ at the $k$-th layer, and is typically a permutation-invariant operator, such as sum, mean, max, or attention-weighted sum~\cite{vaswani2017attention}. $\operatorname{CO}^{(k)}$ maps the aggregated message to higher-level representations at the $k$-th layer, and is usually implemented as a learnable function, such as a multilayer perceptron or a gated unit~\cite{cho2014learning}.

Such a message-passing mechanism has achieved great success in modeling graph-structured data and is highly adaptable to large-scale settings with complex user relationships, ranging from high-frequency financial trading~\cite{li2024anomaly} to industrial recommender systems~\cite{ko2022survey}. 

\subsection{Graph Federated Learning}\label{sec:pre:fl}

Although GNNs have shown strong effectiveness in processing large volumes of data, message passing typically requires collecting data on a central server, which raises data-sharing and privacy concerns~\cite{voigt2017eu}. Therefore, GNNs have recently been explored in the context of federated learning~\cite{li2020review}, leading to graph federated learning (GFL)~\cite{zhou2025fedtps, yu2025homophily, yuintegrating}. Generally, 
GFL aims to collaboratively train a global model across multiple clients without sharing their raw data, thereby preserving data privacy.

Formally, we consider a federated setup with a central server $S$ and a set of $M$ clients. Each client $i \in \{1, 2, \cdots, M\}$ possesses its own local graph data $\mathcal{G}_i = (\mathcal{V}_i, \mathcal{E}_i)$, where $\mathcal{V}_i$ is the node set and $\mathcal{E}_i$ is the edge set, respectively. In addition, we denote the node feature matrix as $\mathbf{X}_i$ and label matrix as $\mathbf{Y}_i$. The goal is to collaboratively train a global GNN on the central server without exposing any local data on clients. Let \(\mathbf{w}\) denote the GNN parameters. The objective is to achieve their optimal values by minimizing the empirical risk over all clients, i.e.,
\begin{equation}\label{eq:fl}
\begin{aligned}
    \hat{\mathbf{w}}^*& = \arg \min_{\mathbf{w}} \hat{\mathcal{L}}(\mathbf{w};\{\mathcal{G}_i, \mathbf{X}_i, \mathbf{Y}_i\}_{i=1}^M)\\ 
    &= \sum_{i=1}^M p_i \hat{\mathcal{L}}_i(\mathbf{w}; \mathcal{G}_i, \mathbf{X}_i, \mathbf{Y}_i),
\end{aligned}
\end{equation}
where $p_i$ is the aggregation weight, typically defined to reflect the relative importance or data size of client $i$, and $\hat{\mathcal{L}}_i(\mathbf{w};\mathcal{G}_i, \mathbf{X}_i, \mathbf{Y}_i)$ is the local empirical risk on client $i$, typically defined as the cross-entropy loss. 

To approximate the search for $\hat{\mathbf{w}}^*$ without sharing data with the central server, we typically follow the federated optimization paradigm of FedAvg~\cite{mcmahan2017communication}. 
FedAvg involves multiple rounds in which clients perform local training on their private data and share updated model parameters with a central server, which aggregates them into a global model.
Specifically, in each round $t$, the server broadcasts the global model $\mathbf{w}^{(t)}$ to all clients. Each client $i$ performs local updates to obtain $\mathbf{w}_i^{(t+1)}$, and the server then aggregates these updates to produce the next global model, following $\mathbf{w}^{(t+1)} = \sum_{i=1}^M p_i \mathbf{w}_i^{(t+1)}$. This procedure will be  repeated for 
$T$ communication rounds.

\subsection{Right To Be Forgotten}
\label{sec:pre:rtbf}
GFL is better aligned with data-sharing regulations than its centralized counterparts, as data are processed locally on clients and never directly shared with the central server. However, with increasing attention to privacy, new requirements must be fulfilled, such as the ``\emph{right to be forgotten}" (RTBF)~\cite{voigt2017eu,bonta2022california}. 

In the literature of GFL, RTBF refers to the requirement that when a user withdraws from the system, the central server and all other clients should remove the message learned from that user. This helps better preserve user privacy and prevents malicious clients from misusing such information. 
Formally, consider a trained global model $\hat{\mathbf{w}}^*$ that approximates the solution of Eq.~\eqref{eq:fl} and suppose the $j$-th client with local graph data $\mathcal{G}_j$ requests to be unlearned. Let $\operatorname{EXT}_j(\mathbf{w};\mathcal{G}, \mathbf{X}, \mathbf{Y})$ denote an extraction function that measures the degree of information about $\mathcal{G}_j$ that can be inferred directly from $\mathbf{w}$ and indirectly from a set of local graph data $(\mathcal{G}, \mathbf{X}, \mathbf{Y})$. The goal of RTBF in GFL is to update the original model parameters from \(\hat{\mathbf{w}}^*\) to \(\hat{\mathbf{w}}^{\mathrm{u}}\) such that
\begin{equation}
\begin{aligned}
\operatorname{EXT}_j\big(\hat{\mathbf{w}}^{\mathrm{u}}; \{\mathcal{G}_i,\mathbf{X}_i,&  \mathbf{Y}_i\}_{i=1}^M\backslash\{\mathcal{G}_j,\mathbf{X}_j, \mathbf{Y}_j\}\big) \\  \ll &\operatorname{EXT}_j\big(\hat{\mathbf{w}}^*; \{\mathcal{G}_i,\mathbf{X}_i, \mathbf{Y}_i\}_{i=1}^M\big),
\end{aligned}
\end{equation}
i.e., the message about $(\mathcal{G}_i, \mathbf{X}_i, \mathbf{Y}_i)$ that can be extracted is notably reduced. This conceptual definition is adapted from~\cite{golatkar2020eternal}, further adjusted for GFL.

At first glance, it may seems that GFL is well suited for RTBF, as we can straightforwardly remove a client once the corresponding user raises a removal request. However, during federated learning, the user information has already been encoded into model parameters, shared with the central server via FedAvg as in Eq.~\eqref{eq:fl}, and then distributed to other clients. By membership inference attacks~\cite{liu2021machine} or model inversion attacks~\cite{he2019model}, the user message may still be at risk of being extracted by others, even after the user has withdrawn from the FGL system.

\subsection{Machine Unlearning} 
\emph{Machine unlearning} (MU)~\cite{cao2015towards} has been widely explored and has shown great promise for fulfilling the primary goal of RTBF. MU is first studied in the context of differential privacy~\cite{abadi2016deep}, where the goal is to ensure that the output distribution of the unlearned model is nearly indistinguishable from that of a model trained as if the unlearned data have never been included, up to a small privacy hyper-parameter. This classical DP definition has led to many elegant MU methods. 
For example, \citet{golatkar2020eternal} study how to add Gaussian noise to parameters that balances between privacy preservation and performance preservation, which is given by 
\begin{equation}
    \hat{\mathbf{w}}^{\mathrm{u}}=\hat{\mathbf{w}}^* + (\lambda_0\sigma^2)^{1/4} \mathbf{H}_{\rm r}^{-1/4}n,~\text{where}~n\sim\mathcal{N}(0,1),
\end{equation}
with $\lambda_0$ a trade-off hyper-parameter, $\sigma$ a prior  variance, and $H_{\rm r}$ the Hessian matrix with respect to 
$$\hat{\mathcal{L}}(\mathbf{w};\{\mathcal{G}_i, \mathbf{X}_i,  \mathbf{Y}_i\}_{i=1}^M\backslash\{\mathcal{G}_j, \mathbf{X}_j,  \mathbf{Y}_j\})$$. 
Overall, the fractional Hessian inverse $H_{\rm r}^{-1/4}$ shapes the noise to be aligned with the local curvature,  mitigating performance degradation while achieving the desired level of privacy. \citet{guo2019certified} certifies that natural gradient updates, following
\begin{equation}
    \mathbf{w}^{(t+1)}=\mathbf{w}^{(t)}+\mathbf{H}_{\rm u}^{-1}\nabla_{\mathbf{w}}\hat{\mathcal{L}}(\mathbf{w};\mathcal{G}_j,\mathbf{X}_j,  \mathbf{Y}_j),
\end{equation}
at each unlearning step $t$, 
where $\mathbf{H}_{\rm u}$ is the Hessian matrix for $\hat{\mathcal{L}}(\mathbf{w};\mathcal{G}_j,\mathbf{X}_j,  \mathbf{Y}_j)$. Generally, DP-based unlearning methods enjoy relatively strong theoretical guarantees. However, on the other side, they typically rely on linear assumptions, require tedious hyper-parameter tuning, and often cause substantial degradation in overall performance.

Recent research offers a more practical definition of MU, directly increasing the risk $\hat{\mathcal{L}}(\mathbf{w};\mathcal{G}_j,\mathbf{X}_j,  \mathbf{Y}_j)$ to be unlearned, leading to \emph{gradient ascent} (GA)-based MU methods~\cite{fan2023salun,wang2025rethinking}. Beyond an explicit unlearning objective, GA-based methods further preserve the model performance on other data that are not intended to be unlearned, leading to a bi-objective as
\begin{equation}
\begin{aligned}
    \arg \min_{\mathbf{w}} - \hat{\mathcal{L}} & (\mathbf{w};\mathcal{G}_j,\mathbf{X}_j,  \mathbf{Y}_j)\\+& \lambda\hat{\mathcal{L}}(\mathbf{w};\{\mathcal{G}_i,\mathbf{X}_i,  \mathbf{Y}_i\}_{i=1}^M\backslash\{\mathcal{G}_j,\mathbf{X}_j,  \mathbf{Y}_j\}),\label{eq:ga}    
\end{aligned}
\end{equation}
where $\lambda$ is a trade-off hyper-parameter.
Although less theoretically grounded, GA-based methods have been shown to achieve more efficient and reliable MU effects than DP-based methods in practice. However, since $\hat{\mathcal{L}}$ is unbounded above, maximizing $\hat{\mathcal{L}}(\mathbf{w};\mathcal{G}_j,\mathbf{X}_j,  \mathbf{Y}_j)$, or equivalently minimizing $-\hat{\mathcal{L}}(\mathbf{w};\mathcal{G}_j, \mathbf{X}_j,  \mathbf{Y}_j)$, can lead to gradient explosion~\cite{wang2025rethinking}, thereby degrading performance. 

This problem can be even more severe for GFL. Although raw data are not shared across clients, information can still propagate between them through the exchanged model parameters~\cite{NEURIPS2021_34adeb8e}. Such implicit message passing means that the backpropagation of unlearning gradients may influence not only the data to be unlearned but also a larger portion of the remaining data than in non-graph models. Moreover, since GNNs rely heavily on the underlying graph structure, aggressively unlearning specific nodes, edges, or subgraphs can distort the learned topological representations, potentially harming the overall quality of the global model. Therefore, we need further improve upon existing  GA-based unlearning methods to make it more suitable for GFL scenarios.

\section{Our Proposed Method}
In this section, we propose \emph{\underline{\textbf{Fed}}erated unlearning method with \underline{\textbf{G}}radient \underline{\textbf{C}}orrection and \underline{\textbf{V}}irtual client} (FedGCV), an enhanced GA-based MU method specifically designed for GFL. FedGCV makes two primary contributions. First, we introduce a \textbf{gradient correction} strategy that explicitly addresses the unlearning-preservation trade-off. Second, we propose a \textbf{virtual client} strategy that aims to preserve the original topological structure. Together, these components yield a reliable RTBF framework that better preserves overall model performance than conventional GA-based methods such as Eq.~\eqref{eq:ga} for GFL.

\subsection{Gradient Correction} 
Recall that conventional GA-based methods formulate MU as a bi-objective task, where we simultaneously maximize the risk on the target data, i.e., \(\max_{\mathbf{w}} \hat{\mathcal{L}}(\mathbf{w}; \mathcal{G}_j,\mathbf{X}_j,  \mathbf{Y}_j)\), and minimize the risk on the remaining data, i.e., \(\min_{\mathbf{w}} \hat{\mathcal{L}}(\mathbf{w}; \{\mathcal{G}_i,\mathbf{X}_i,  \mathbf{Y}_i\}_{i=1}^M \setminus \{\mathcal{G}_j,\mathbf{X}_j,  \mathbf{Y}_j\})\). However, previous work has shown that these two objectives are inherently conflicting~\cite{wang2025gru}, indicating that substantially increasing \(\hat{\mathcal{L}}(\mathbf{w}; \mathcal{G}_j,\mathbf{X}_j,  \mathbf{Y}_j)\) will typically hinder progress in reducing \(\hat{\mathcal{L}}(\mathbf{w}; \{\mathcal{G}_i,\mathbf{X}_i,  \mathbf{Y}_i\}_{i=1}^M \setminus \{\mathcal{G}_j,\mathbf{X}_j,  \mathbf{Y}_j\})\). Further tuning $\lambda$ only amplifies this trade-off between unlearning and retention, resulting either in less effective unlearning or in more severe degradation of model performance. Therefore, in this section, we explore an alternative strategy, beyond the bi-objective formulation in Eq.~\eqref{eq:ga}, that can effectively mitigate this trade-off.



Our key motivation is to ensure that each unlearning update has minimal negative impact on  the performance of other normal data, leading to a constrained optimization problem 
\begin{equation}\label{eq7}
\begin{aligned}
    & \quad~~\arg \min_{\mathbf{w}} - \hat{\mathcal{L}}  (\mathbf{w};\mathcal{G}_j,\mathbf{X}_j,  \mathbf{Y}_j)\\
    \text{s.t.}~&
    \hat{\mathcal{L}}(\mathbf{w};\{\mathcal{G}_i,\mathbf{X}_i,  \mathbf{Y}_i\}_{i=1}^M\backslash\{\mathcal{G}_j,\mathbf{X}_j,  \mathbf{Y}_j\})\\\le&\hat{\mathcal{L}}(\hat{\mathbf{w}}^*;\{\mathcal{G}_i,\mathbf{X}_i,  \mathbf{Y}_i\}_{i=1}^M\backslash\{\mathcal{G}_j,\mathbf{X}_j,  \mathbf{Y}_j\}).
\end{aligned}
\end{equation}
However, Eq.~\eqref{eq7} is hard to be solved. Therefore, following~\cite{lopez2017gradient}, we approximate the constraint in Eq.~\eqref{eq:ga} under a first-order assumption as follows:
\begin{equation}
\begin{aligned}
    \langle \mathbf{g}^{(t)}_{\rm u},&\mathbf{g}^{(t)}_{\rm r}\rangle  :=\big\langle \nabla_{\mathbf{w}}{\partial\hat{\mathcal{L}}(\mathbf{w}^{(t)};\mathcal{G}_j)},\\ & \nabla_{\mathbf{w}}{\partial\hat{\mathcal{L}}(\mathbf{w}^{(t)};\{\mathcal{G}_i,\mathbf{X}_i,  \mathbf{Y}_i\}_{i=1}^M\backslash\{\mathcal{G}_j,\mathbf{X}_j,  \mathbf{Y}_j\})}\big\rangle\geq 0,
\end{aligned}    \label{inequality}
\end{equation}
at each updating step $t$.
Therein, $\mathbf{g}_{\rm u}$ denotes the direction of unlearning updates for the $j$-th client, and $\mathbf{g}_{\rm r}$ denotes the direction of parameter updates for other clients. \citet{wang2025gru} further study this constrained optimization and find the optimal solution as  
\begin{equation}
    \hat{\mathbf{g}}_{\rm u}^{(t)}=\mathbf{g}_{\rm u}^{(t)}-\frac{\mathrm{min}(\langle \mathbf{g}_{\rm u}^{(t)},\mathbf{g}_{\rm r}^{(t)}\rangle,0)}{||\mathbf{g}_{\rm r}^{(t)}||^2}\cdot \mathbf{g}_{\rm r}^{(t)},\label{gru}
\end{equation}
It includes two scenarios. If the constraint in Eq.~\eqref{inequality} is satisfied, $\langle \mathbf{g}_{\rm u}^{(t)},\mathbf{g}_{\rm r}^{(t)}\rangle\geq 0$ and there is no need to correct unlearning gradients. On the other side, if $\langle \mathbf{g}_{\rm u}^{(t)},\mathbf{g}_{\rm u}^{(t)}\rangle<0$, $\mathbf{g}_{\rm u}$ will be projected into the orthogonal direction of $\mathbf{g}_{\rm r}$ to obtained the corrected gradient direction $\hat{\mathbf{g}}_{\rm u}$.

\begin{figure}[]
 \centering
 \subfloat{\includegraphics[width=0.6\columnwidth]{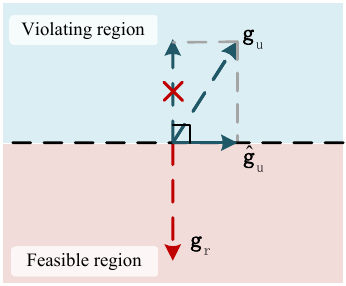}}\label{GRU-image}
 \caption{Illustration of the gradient correction rule.}
 \vskip -0.2in
 \label{visio-GRU}
\end{figure}

In the GFL literature, the above discussion leads to a gradient correction rule that can be coordinated by the central server, which makes it well suited to FedAvg. Specifically, when the central server collects updates from all clients at step $t$, we can compute the residuals $\boldsymbol\Delta_{\rm u}$ and $\boldsymbol\Delta_{\rm r}$ with respect to the unlearning client and all the other clients that should be preserved, which is given by
\begin{equation}
    \boldsymbol\Delta^{(t+1)}_{\rm u}=\mathbf{w}^{(t)} -p_j\mathbf{w}_j^{(t+1)},
\end{equation}
and
\begin{equation}
    \boldsymbol\Delta^{(t+1)}_{\rm r}=\sum_{i=1;i\ne j}^M p_i \mathbf{w}_j^{(t+1)}-\mathbf{w}^{(t)},
\end{equation}
respectively. Then, the aggregation procedure with unlearning can be written as
\begin{equation}
\begin{aligned}
\mathbf{w}^{(t+1)}=\mathbf{w}^{(t)}&+ \boldsymbol\Delta^{(t+1)}_{\rm u}-\\&\frac{\mathrm{min}(\langle \boldsymbol\Delta^{(t+1)}_{\rm u},\boldsymbol\Delta^{(t+1)}_{\rm r}\rangle,0)}{||\boldsymbol\Delta^{(t+1)}_{\rm r}||^2}\cdot \boldsymbol\Delta^{(t+1)}_{\rm r}.
\end{aligned}
\end{equation}
This formulation adapts FedAvg to the RTBF setting, erasing the information from the client to be unlearned while, to a large extent, preserving the overall model performance.

\subsection{Virtual Client}
After unlearning the subgraph data of the specific client that needs to be forgotten, since both topological structures and node attributes of the unlearned client are lost, the performance of the global model will be notably dropped. Therefore, in the context of FedGCV, relying solely on gradient correction may lead to a decline in global model utility. To restore federated learning performance without revealing the subgraph data of the unlearned client, we propose a novel mechanism termed virtual client, where we synthesize graphs that are similar to that of the unlearned client and maintain them by the central server. Specifically, we first extract spectral information and node feature distributions from the real subgraph data. We then employ a \emph{variational graph autoencoder} (VGAE)~\cite{kipf2016variational} to generate synthetic subgraphs with similar structural properties and feature distributions. These synthetic graphs are used to replace unlearned clients in subsequent federated training. This achieves adaptive collaborative training that restores utility without privacy rebound.

\subsubsection{Extraction of spectral information}\label{SpectralStatistical}
To construct virtual clients, we employ a VGAE to capture the latent distribution of the unlearned client. Moreover, to reconstruct the topological structures of the unlearned client, we further employ spectral constraints to guide the generation of edges between nodes across clients.

Let the subgraph on a forgotten client be $\mathcal{G}_j=(\mathcal{V}_j,\mathcal{E}_j)$, $n=|\mathcal{V}_j|$, the adjacency matrix be $\mathbf{A}_j\in\mathbb{R}^{n\times n}$, and the node feature matrix be $\mathbf{X}_j\in\mathbb{R}^{n\times d}$. First, we employ propagation matrices, which are consistent with the message-passing mechanism:
\begin{equation}
\tilde{\mathbf{A}}_j=\mathbf{A}_j+\mathbf{I},\quad
\hat{\mathbf{A}}_j = \tilde{\mathbf{D}}_j^{-\frac12}\tilde{\mathbf{A}}_j\tilde{\mathbf{D}}_j^{-\frac12}.
\end{equation}

To capture the structural topologies of forgotten subgraphs, we then extract information from their Laplacian matrices. The graph Laplacian matrix $\mathbf{L}_j$ of subgraph $\mathcal{G}_j$ is defined as $\mathbf{L}_j=\mathbf{D}_j-\mathbf{A}_j$, where $\mathbf{D}_j$ is the degree matrix of $\mathcal{G}_j$ and $\mathbf{A}_j$ is the adjacency matrix of $\mathcal{G}_j$. After performing normalization on the Laplace matrix, we can obtain the normalized Laplacian matrix $\mathbf{L}_j^\text{norm}$ as:
\begin{equation}\label{lj_norm}
    \mathbf{L}_j^\text{norm}=\mathbf{I}-\mathbf{D}_j^{-\frac12}\mathbf{A}_j\mathbf{D}_j^{-\frac12}.
\end{equation}
Eq.~\eqref{lj_norm} eliminates the influence of the degree of the Laplacian matrix, which further ensures the graph spectrum is bounded and yields more stable numerical results.

Since $\mathbf{L}_j^\text{norm}$ is a semi-definite matrix, we can perform an eigenvalue decomposition on $\mathbf{L}_j^\text{norm}$ to obtain its eigenvalues and eigenvectors. The equation can be expressed as $\mathbf{L}_j^\text{norm} = \mathbf{U}_j\mathbf{\Lambda}_j\mathbf{U}_j^\top$, and take the first $k$ smallest eigenvalues (low-frequency component) $\mathbf{\Lambda}_{jk}$ and their corresponding eigenvectors $\mathbf{U}_{jk}$. The low-frequency subspace characterizes the global connectivity and smooth structure of the subgraph, which is crucial for preserving graph spectral information during generation.

\subsubsection{VGAE Modeling}
To capture the node feature distributions of the forgotten graph, VGAE learns latent representations of subgraphs in an unsupervised manner so as to generate a similar but privacy-protective virtual subgraph. 

Although VGAE is capable of capturing node feature distribution, it may not learn the structural topologies of graphs well. To mind this gap, we exploit the extracted spectral properties in~\cref{SpectralStatistical} and integrate them in the synthesis process of virtual graphs. To preserve the spectral characteristics of the forgotten graph in the synthetic graph, we project the latent variables onto the subspace spanned by $\mathbf{U}_k$. The projected latent variable $\mathbf{Z}_\text{proj}$ can be obtained from the following formula:
\begin{equation}\label{z_proj}
    \mathbf{Z}_\text{proj} = \mathbf{U}_k \mathbf{U}_k^\top \mathbf{Z},
\end{equation}
where $\mathbf{Z}$ represents the inferred latent distribution of the forgotten graph. Note that Eq.~\eqref{z_proj} only permits $\mathbf{Z}$ to utilize structures expressible by the original subgraph's spanned subspace. Afterwards, we can obtain the synthetic graphs by performing graph reconstruction based on $\mathbf{Z}_\text{proj}$. This reconstruction process is implemented as an inner product following~\cite{kipf2016variational}. To be specific, this process uses $\mathbf{Z}_\text{proj}$ to obtain the edge probability matrix $\mathbf{P}$:
\begin{equation}
    \mathbf{P}=\sigma(\mathbf{Z}_\text{proj}\mathbf{Z}_\text{proj}^\top).
\end{equation}
Subsequently, we employ a thresholding strategy to generate the synthetic adjacency matrix $\mathbf{A}_\text{syn}$ from $\mathbf{P}$, converting continuous probability into a matrix composed of only 0 and 1. The process can be expressed as follows:
\begin{equation}
    (\mathbf{A}_\text{syn})_{ij}=\mathbb{I}\big[\mathbf{P}_{ij}>\gamma\big],
\end{equation}
where $\mathbb{I}[\cdot]$ is an indicator function, and $\gamma\in(0,1)$ is the probability threshold used to determine whether to generate an edge.

Finally, to generate a virtual subgraph, we still need to integrate features with the mean vector $\boldsymbol{\mu}$ and standard deviation vector $\boldsymbol{\sigma}$ of the original subgraph inferred by VGAE. To avoid leaking the original subgraph data $\mathcal{G}$, we also employ a high-privacy statistical sampling strategy to obtain the feature matrix $\mathbf{X_{syn}}$ reflecting the node feature distribution of the virtual subgraph. Each element $(\mathbf{X_{syn}})_{ij}$ follows a Gaussian distribution that satisfies the mean and standard deviation of the original subgraph:
\begin{equation}
    (\mathbf{X}_\text{syn})_{ij}\sim \mathcal{N}\big(\boldsymbol{\mu}, \mathrm{diag}(\boldsymbol{\sigma}^2)\big).
\end{equation}

Only the node-specific distributions are retained without directly using the original data $\mathcal{G}$, which significantly preserves the privacy of the forgotten graph. The generated $\mathbf{A}_\text{syn}$ incorporates the structure of the synthetic subgraph, describing the topological structure of the virtual subgraph, and the generated $\mathbf{X}_\text{syn}$ contains the node attributes and features of the synthetic subgraph, describing the node content. Both $\mathbf{A}_\text{syn}$ and $\mathbf{X}_\text{syn}$ can collectively represent an attributed graph $\mathcal{G}_\text{virtual}$ usable in GFL, enabling performance compensation functionality in subsequent parameter updates.

\section{Experiments}
In this section, we evaluate our proposed methodological framework and compare it against state-of-the-art federated unlearning methods on four widely used graph datasets.

\subsection{Experimental Setups}
We begin by detailing the experimental setups adopted in this paper, including datasets, evaluation protocols, implementation settings, and baseline methods.

\textbf{Datasets.} Each client possesses a subgraph that is a subset of the original large graph. To be specific, we utilize four real-world datasets that are widely employed in graph learning tasks, including Cora, CiteSeer~\cite{yang2016Plato}, PubMed~\cite{yang2016Plato}, and Tolokers~\cite{platonov2023critical}. Following prior works~\cite{yu2025homophily}, we simulate the real GFU scenario by partitioning the original graph of the mentioned datasets with the METIS algorithm~\cite{karypis1998fast}. Detailed information about these datasets is provided in Appendix~\ref{datasets}.

\label{mia-eva}
\textbf{Evaluation Protocols.} Our evaluation metrics align with our task motivation for graph federated unlearning. That is, maintaining both high unlearning efficiency and overall utility. (1) \emph{Membership inference attack} (MIA)~\cite{sablayrolles2019whiteboxvsblackboxbayes}: MIAs evaluate privacy leakage and unlearning efficiency by measuring whether an adversary can distinguish training samples from non-members based on the model outputs. If the MIA rate is lower, the unlearning efficiency and privacy protection is better. Under this paradigm, we employ MAIT as an evaluation metric. MAIT determines whether a sample is a training member by leveraging the difference between its loss and gradient evolution track during model training, and evaluates the degree of unlearning based on the decline in this distinguishability. For a detailed implementation process, please see Appendix~\ref{MIA evaluation}. (2) \emph{Model accuracy} (MA): After the entire unlearning process concludes, we will measure the global model accuracy to assess its overall performance. If the MA is higher, the performance retention is more reliable.

\begin{table*}[t]
\centering
\caption{Overall performance evaluation of graph federated unlearning. The percentage next to the dataset name indicates the model accuracy achieved by FedAvg without unlearning.}
\label{tab:overall-gfu}
\renewcommand{\arraystretch}{1.15}
\setlength{\tabcolsep}{7pt}
\begin{tabular}{lll|ll}
\hline
\multicolumn{1}{c}{}      & \multicolumn{2}{c|}{Cora (82.42\%)}                                       & \multicolumn{2}{c}{CiteSeer (70.14\%)}                                   \\ \hline
\multicolumn{1}{c}{Methods} & \multicolumn{1}{c}{Performance $\uparrow$} & \multicolumn{1}{c|}{MIA Rate $\downarrow$}  & \multicolumn{1}{c}{Performance $\uparrow$} & \multicolumn{1}{c}{MIA Rate $\downarrow$}  \\ \hline
retrain                   & 68.67\%                                 & \textbf{3.70\%}                & 61.83\%                                 & \textbf{0.00\%}                        \\
PGA~\cite{halimi2023federatedunlearningefficientlyerase}                       & 71.91\%                                 & 12.96\%                        & 56.91\%                                 & 12.70\%                       \\
SGA-Random~\cite{Bottou2010LargeScaleML}                & 68.09\%                                 & 14.81\%                        & 48.88\%                                 & 14.29\%                       \\
SGA-Degree                & 68.80\%                                 & 12.22\% & 51.82\%                                 & 9.52\%                        \\
EWC-SGA~\cite{9964015}                   & 66.89\%                                 & 9.26\%                         & 42.97\%                                 & 22.22\%                       \\
Noisy-GD~\cite{chourasia2023forgetunlearningtruedatadeletion}                  & \underline{72.42\%}                                 & 87.04\%                        & 59.62\%                                 & 88.89\%                       \\
ULKD~\cite{Wu2021AFG}                      & 70.79\%                                 & 18.52\%                        & 63.23\%                                 & 15.87\%                       \\
ReGEnUnlearn~\cite{liu2025subgraph}              & 72.04\%                                 & 9.09\%                         & \underline{63.25\%}                                 & 12.39\%                       \\
\textbf{FedGCV (Ours)}      & \textbf{79.41\%}                        & \underline{5.56\%}                         & \textbf{69.07\%}                                 & \underline{7.94\%}                        \\ \hline
                          & \multicolumn{2}{c|}{PubMed (86.52\%)}                                     & \multicolumn{2}{c}{Tolokers (78.10\%)}                                   \\ \hline
Methods                     & \multicolumn{1}{c}{Performance $\uparrow$} & \multicolumn{1}{c|}{MIA Rate $\downarrow$}  & \multicolumn{1}{c}{Performance $\uparrow$} & \multicolumn{1}{c}{MIA Rate $\downarrow$}  \\ \hline
retrain                   & \textbf{85.73\%}                        & \textbf{0.52\%}                & \textbf{76.58\%}                        & 2.89\%                        \\
PGA~\cite{halimi2023federatedunlearningefficientlyerase}                       & 76.66\%                                 & 6.81\%                         & 65.65\%                                 & 11.30\%                       \\
SGA-Random~\cite{Bottou2010LargeScaleML}                &  75.62\%          &  10.65\% &  67.77\%          &  5.34\% \\
SGA-Degree                &   75.70\%          &   9.69\%  &   65.74\%          &   9.46\% \\
EWC-SGA~\cite{9964015}                   &   77.07\%          &   10.47\% &   67.55\%          &   7.64\% \\
Noisy-GD~\cite{chourasia2023forgetunlearningtruedatadeletion}                  &   80.67\%          &   7.07\%  &   66.08\%          &   1.85\% \\
ULKD~\cite{Wu2021AFG}                      & 82.54\%                                 & 8.12\%                         & 62.57\%                                 & 10.96\%                       \\
ReGEnUnlearn~\cite{liu2025subgraph}              & 77.41\%                                 & 7.92\%                         & 67.47\%                                 & \underline{1.03\%}                        \\
\textbf{FedGCV (Ours)}      & \underline{83.35\%}                                 & \underline{3.66\%}                         & \underline{69.17\%}                                 & \textbf{0.18\%}               \\ \hline
\end{tabular}
\end{table*}

\textbf{Implementation Settings.}
We set the random seed to $s=2025$ and train a $2$-layer GCN with hidden dimension $d=64$, dropout rate $p=0.5$, learning rate $\eta=10^{-2}$, weight decay $\lambda=5\times 10^{-4}$, and batch size $B=128$ for federated training. We run federated learning for $R=30$ communication rounds over $K=10$ clients with full participation ratio $\rho=1.0$ (i.e., all clients are selected in each round).
For interactive unlearning, we optimize the unlearning model for $E_u=30$ epochs with learning rate $\eta_u=2\times 10^{-2}$ and dropout $p_u=0.3$. We adopt the NPO-based unlearning loss~\cite{zhang2024negative}, an advanced GA-based method, with temperature parameter $\beta=5.0$ and scale the corrected unlearning update by $s_f=50$. To reduce catastrophic drift during unlearning, we clip the per-step update norm by $c_{\max}=10.0$ and project the parameter drift to satisfy $\lVert\boldsymbol{\theta}-\boldsymbol{\theta}_0\rVert_2\le \tau$, where $\tau=10.0$.
For Graph-MIA–aware training, we additionally enforce a fixed-threshold alignment by pushing the target loss above the fixed-pre threshold by a margin $m=0.5$ with weight $\lambda_m=3.0$. For post-unlearning virtual repair, we conduct $R_v=5$ extra federated rounds. The adjacency reconstruction uses a threshold $\tau_a=0.7$, and feature perturbation uses noise standard deviation $\sigma_x=0.1$.

\textbf{Baselines.} We consider the following joint  federated unlearning methods:
(1) Retraining from scratch: This method retrains an initialized model using data from the retaining clients, representing the ideal state that unlearning tasks aim to approximate. (2) Projected Gradient Ascent(PGA)~\cite{halimi2023federatedunlearningefficientlyerase}: Projection Gradient Ascent (PGA): PGA is an unlearning method designed to maximize local client loss within a controlled range. (3) Elastic Weight Consolidation-Stochastic Gradient Ascent (EWC-SGA)~\cite{9964015}: EWC-SGA combines stochastic gradient ascent on the removed client loss with elastic weight consolidation to restrict updates on parameters that are important to the remaining clients to remove target clients without requiring full model retraining. (4) Noisy Gradient Descent (Noisy-GD)~\cite{chourasia2023forgetunlearningtruedatadeletion}: Noisy-GD injects random noise into gradient descent updates to reduce the influence of removed data by perturbing model parameters. (5) ULKD~\cite{Wu2021AFG}: ULKD is a server-side storage history method that uses distillation techniques to eliminate target client data-sharing while enhancing model utility. (6) SGA-Random~\cite{Bottou2010LargeScaleML}: SGA-Random performs stochastic gradient ascent using randomly sampled data from the removed client to erase its contribution from the model. (7) SGA-Degree: SGA-Degree applies stochastic gradient ascent with degree-aware sampling, prioritizing nodes with higher structural influence during unlearning. (8) ReGEnUnlearn~\cite{liu2025subgraph}: ReGEnUnlearn optimizes sampling strategies that minimize cross-client subgraph interference through Reinforcement learning-driven Federated Policy Sampler, while integrating a Parameter-free Graph Prompt Knowledge Distillation module to achieve precise retention of target client's unique structural knowledge and gradient-ascending complete unlearning.


\subsection{Main Results}\label{mainresult}
\cref{tab:overall-gfu} presents the comprehensive results of our proposed method and corresponding baseline methods under two metrics. We draw the following conclusions: Compared to all baseline models, our proposed FedGCV achieves the optimal performance across four datasets (except in the ideal retrain scenario), and in many cases surpasses the performance of retrain. Notably, FedGCV achieves over 90\% unlearning efficiency on the Cora, CiteSeer, PubMed, and Tolokers datasets, second only to the unlearning MIA rate of retraining. Furthermore, our framework compensates for model performance loss to over 90\% of the pre-unlearning accuracy, making it the best among all baseline methods. Even on heterophilic subgraph dataset like Tolokers, it still performs exceptionally well. Compared with the state-of-the-art approach, ReGEnUnlearn~\cite{liu2025subgraph}, our proposed method simultaneously improves utility by 5.21\% and reduces privacy leakage by 2.00\%, on average across four datasets, demonstrating a favorable balance between performance and privacy.

\subsection{Ablation Studies}
\begin{figure}[t]
 \centering
 \subfloat[\footnotesize{Cora}]{\includegraphics[width=0.5\columnwidth]{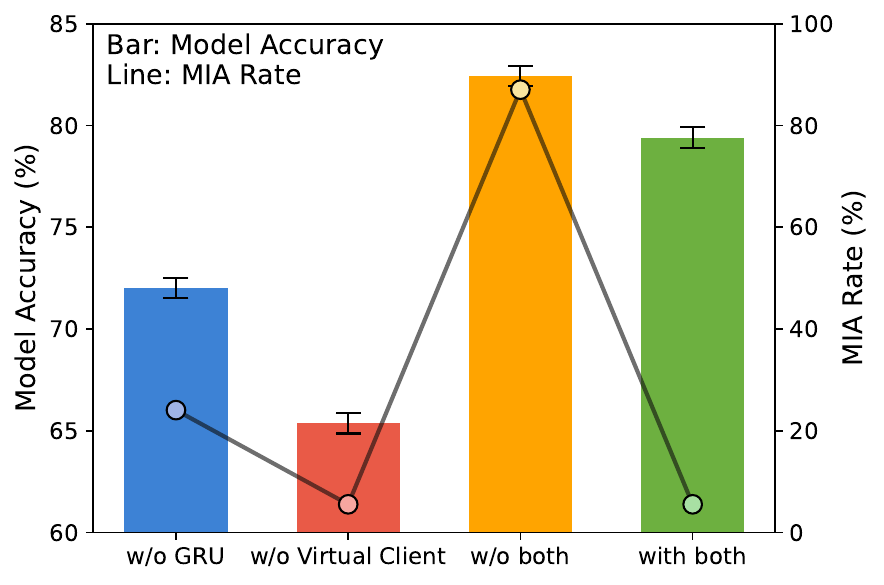}\label{ablation_1}}
 \hfill
 \subfloat[\footnotesize{CiteSeer}]{\includegraphics[width=0.5\columnwidth]{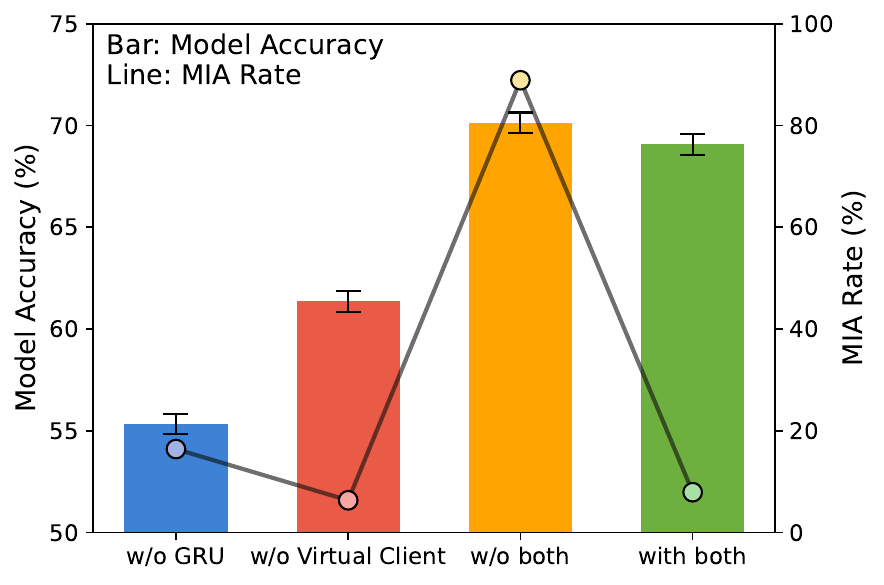}\label{ablation_2}}
 \hfill
  \subfloat[\footnotesize{PubMed}]{\includegraphics[width=0.5\columnwidth]{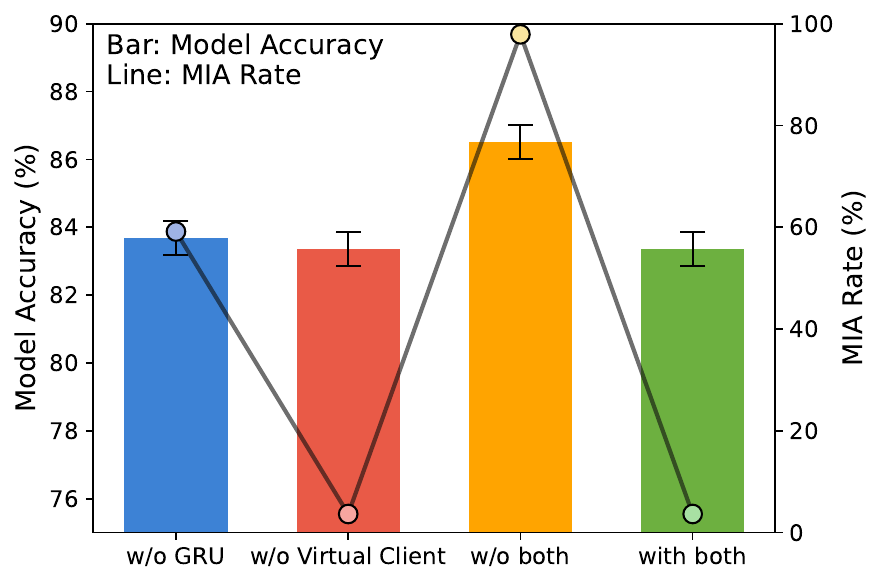}\label{ablation_3}}
 \hfill
  \subfloat[\footnotesize{Tolokers}]{\includegraphics[width=0.5\columnwidth]{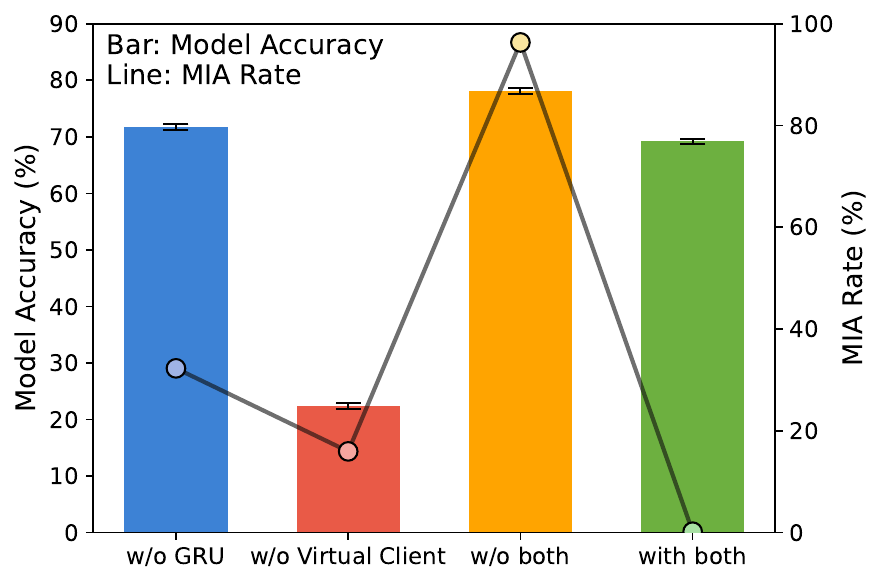}\label{ablation_4}}
 \caption{Unlearning evaluation and model accuracy are measured using four datasets: Cora, CiteSeer, PubMed, and Tolokers. The bar charts represent Model Accuracy, where higher performance is better; the line charts represent the Mia Rate, where lower values indicate higher unlearning levels.}
 \vskip -0.2in
 \label{ablation}
\end{figure}

To evaluate the effectiveness of each module in FedGCV, we conducted ablation studies across four datasets (e.g., Cora, CiteSeer, PubMed and Tolokers) and employed two evaluation metrics. Specifically, we denote simplified approaches achieved by removing ``GRU-based gradient correction for unlearning'' and ``virtual client compensation for model utility'' as ``w/o GRU'' and ``w/o Virtual Client'', respectively. It can be observed in Figure~\ref{ablation} that the overall performance declines whenever any component is removed, indicating each component makes significant contributions. For instance, when the GRU was disabled and replaced with a traditional gradient ascent strategy in models trained on the Cora dataset, the MIA rate reached 24.07\%, an increase of nearly 20\% compared to when it was enabled. In addition, when the Virtual Client was disabled, model accuracy dropped from 82.42\% to 65.37\%. However, after incorporating the Virtual Client, accuracy reached 79.41\%, clearly enhancing model utility.

\subsection{Sensitivity Analysis on Hyperparameters}
\begin{figure}[t]
 \centering
 \subfloat[\footnotesize{Cora}]{\includegraphics[width=0.5\columnwidth]{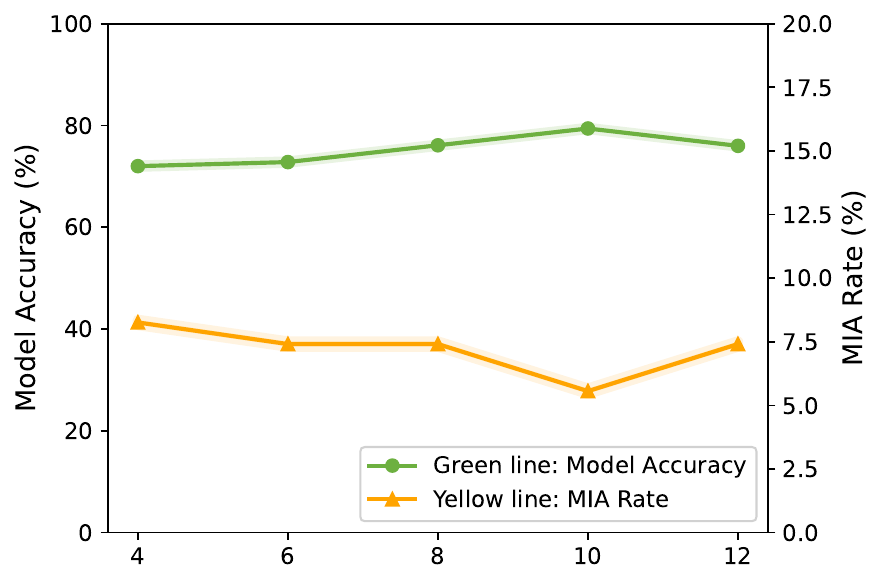}\label{sensitivity_1}}
 \hfill
 \subfloat[\footnotesize{CiteSeer}]{\includegraphics[width=0.5\columnwidth]{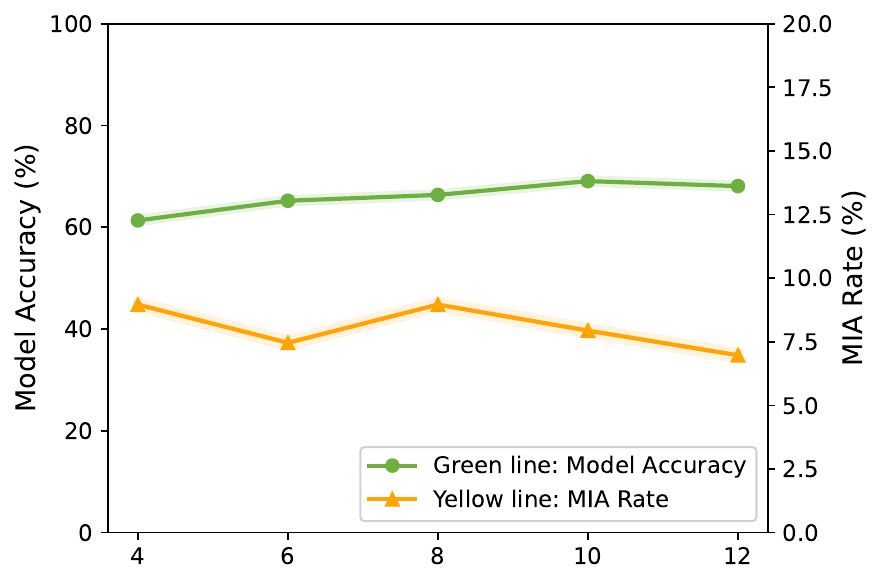}\label{sensitivity_2}}
 \hfill
  \subfloat[\footnotesize{PubMed}]{\includegraphics[width=0.5\columnwidth]{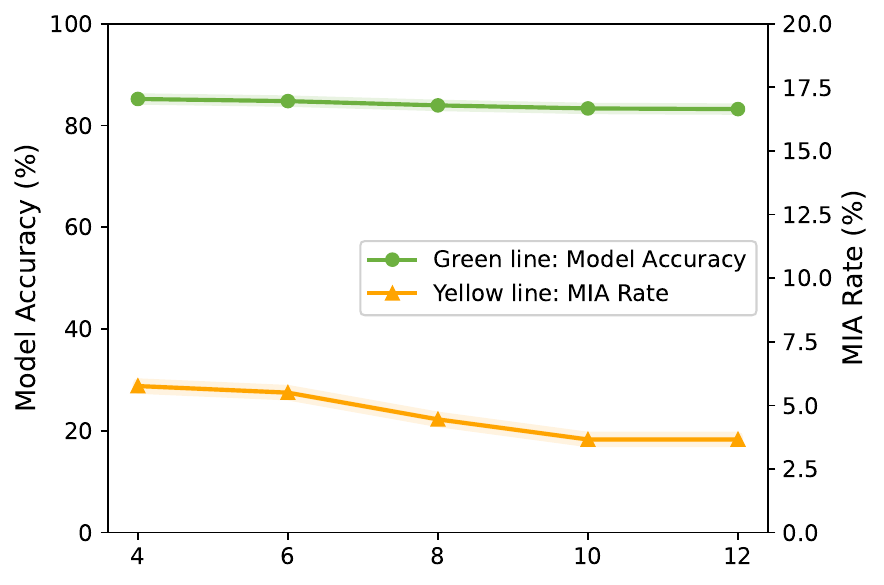}\label{sensitivity_3}}
 \hfill
  \subfloat[\footnotesize{Tolokers}]{\includegraphics[width=0.5\columnwidth]{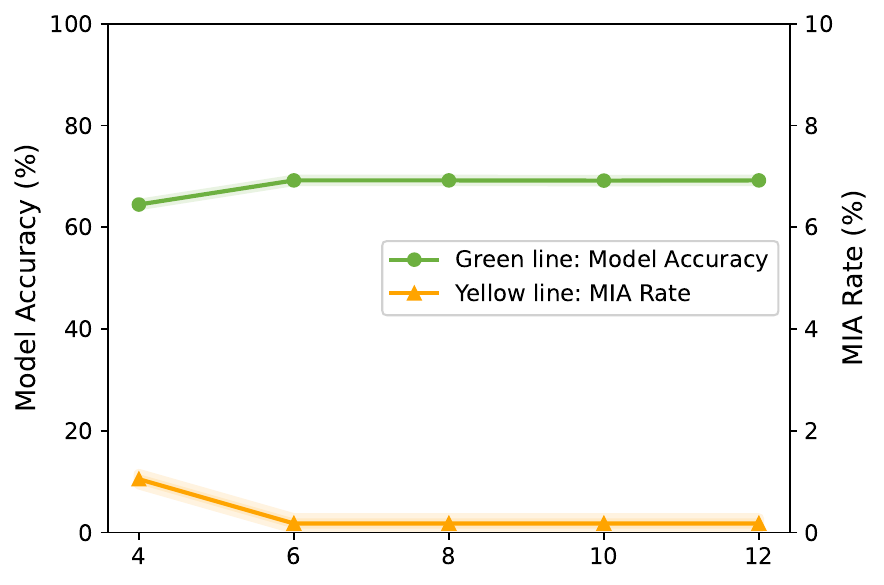}\label{sensitivity_4}}
 \caption{Precision curves, MIA member recognition rates, and their standard deviation bands for different $\tau$ values on the Cora, CiteSeer, PubMed, and Tolokers datasets.}
 \vskip -0.2in
 \label{Sensitivity}
\end{figure}

Here, we conduct a detailed sensitivity analysis of the hyperparameters involved in the proposed FedGCV. Our FedGCV includes a hyperparameter $\tau$ that defines the maximum permissible norm for model parameter updates. Update steps exceeding $\tau$ are scaled proportionally to remain within the $\tau$ threshold. We plotted the MIA rate metric and model accuracy curves across four datasets, demonstrating performance under different hyperparameter values. As shown in Figure~\ref{Sensitivity}, performance fluctuates minimally across varying hyperparameter values, validating that FedGCV exhibits low sensitivity to these parameters as long as they are within reasonable ranges.

\section{Conclusion}
This paper investigates the issue of graph federated unlearning to address data privacy and the right to be forgotten. We propose a novel graph federated unlearning framework, graph \underline{\textbf{Fed}}erated unlearning method with \underline{\textbf{G}}radient \underline{\textbf{C}}orrection and \underline{\textbf{V}}irtual client (FedGCV), which balances model utility while ensuring effective unlearning. Specifically, FedGCV first introduces a gradient rectification unlearning mechanism to correct the direction of forgotten gradients. This minimizes interference with the learning performance of retaining clients while weakening the influence of forgotten clients. Subsequently, we design a VGAE-based virtual client generation strategy to synthesize virtual subgraphs that retain the original client's subgraph node feature distributions and graph spectral information without exposing raw data. These virtual subgraphs maintain similarity to real data in node feature distributions, Laplacian spectral properties, and propagation structures while preserving privacy, enabling effective participation in federated training to compensate for performance loss caused by unlearning. Extensive experimental results demonstrate that FedGCV significantly reduces the risk of member inference attacks while stably maintaining the model's predictive performance on retained data. Compared with various existing federated unlearning methods, FedGCV achieves a superior balance between unlearning effectiveness and model utility, validating its effectiveness and practicality in graph federated unlearning scenarios.

\section*{Impact Statement}
This study advances the development of Graph Federated Unlearning (GFU) methods under the right to be forgotten requirement. These methods simultaneously satisfy privacy protection for clients to be forgotten and the effectiveness demands of the overall model, thereby reducing privacy leakage risks in subgraph learning scenarios. We encourage ongoing research into legally compliant and reliable GFU methods that ensure the sustainability of model training and learning in practical applications while respecting individual rights and intellectual property. By promoting the widespread deployment of GFU methods adaptable to evolving legal and ethical standards, this study contributes to maintaining the trustworthiness of such technologies and their continued generation of social benefits.



\nocite{langley00}

\bibliographystyle{spbasic}      
\bibliography{reference}   

\newpage
\appendix
\onecolumn
\section{Implementation Details}
\subsection{Datasets}\label{datasets}
In this experiment, we employ four representative graph datasets: Cora, CiteSeer, PubMed, and Tolokers, which are widely utilized in graph neural network research.

\textbf{Cora:} The Cora dataset is an academic paper citation network comprising 2,708 nodes, each representing a paper. Each paper's features are represented by a 1,433-dimensional bag-of-words vector, reflecting the presence of specific terms within the paper. The dataset contains 5,429 citation edges, typically processed as an undirected graph. Papers are categorized into 7 topics, including neural networks and genetic algorithms, primarily for node classification tasks.

\textbf{CiteSeer:} The CiteSeer dataset is another academic paper citation network, slightly larger than Cora with 3,327 nodes and 4,732 citation edges. Each paper is represented as a 3,703-dimensional sparse binary vector and categorized into 6 domains, such as artificial intelligence and machine learning. This dataset is frequently used for node classification and graph representation learning tasks, serving as a classic benchmark for graph neural network validation.

\textbf{PubMed:} The PubMed dataset originates from the biomedical literature citation network, comprising 19,717 nodes and 44,338 edges. Each paper's feature is a 500-dimensional TF-IDF vector representing word importance within the paper. There are 3 categories, such as diabetes-related topics. Due to its large scale and sparse real-valued features, PubMed is frequently used for semi-supervised node classification and testing the performance of graph convolutional networks.

\textbf{Tolokers:} The Tolokers dataset originates from a crowdsourcing task platform, where nodes typically represent task participants or tasks themselves. Node counts range from approximately 5,000 to 10,000, with feature dimensions of 300–1,000 sparse vectors representing user or task attributes. This dataset is suitable for evaluating graph neural network performance in real-world scenarios, particularly for federated learning and oblivious learning tasks.

\subsection{MIA Evaluation Metric}\label{MIA evaluation}
In our implementation, MAIT evaluates unlearning via a membership inference attack (MIA) that uses the per-sample loss as the membership statistic, and reports the \textbf{MIA rate} on the target data which should be forgotten. To make the pre/post comparison well-defined, we fix the decision threshold at the value fitted on the pre-unlearning model and reuse it for all post-unlearning evaluations.

Let $\mathcal{D}_{\mathrm{tgt}}$ denote the target dataset.  
Let $\mathcal{D}_{\mathrm{ret}}$ denote a retained dataset drawn from other clients.  
Let $\mathcal{D}_{\mathrm{vir}}$ denote a virtual/non-member dataset generated from $\mathcal{D}_{\mathrm{ret}}$.

Let $\boldsymbol{\theta}$ be the parameters of the model being evaluated, $f(\cdot;\boldsymbol{\theta})$ the model, and $\mathcal{L}(\cdot,\cdot)$ the task loss (cross-entropy).
For each sample $(\mathbf{x},y)$, define the membership statistic as the sample loss:
\[
\ell(\mathbf{x},y;\boldsymbol{\theta}) \triangleq \mathcal{L}\!\left(f(\mathbf{x};\boldsymbol{\theta}),y\right).
\]

Let $\boldsymbol{\theta}^{\mathrm{pre}}$ be the pre-unlearning model. MAIT fits a decision threshold $\tau_{\mathrm{pre}}$ using $\boldsymbol{\theta}^{\mathrm{pre}}$ together with member/non-member reference data. Intuitively, $\tau_{\mathrm{pre}}$ is chosen so that losses of members (from $\mathcal{D}_{\mathrm{tgt}}$) are more likely to fall below the threshold than losses of non-members (from $\mathcal{D}_{\mathrm{vir}}$), enabling membership discrimination.

Let $\mathbb{I}[\cdot]$ be the indicator function.
Given the fixed threshold $\tau_{\mathrm{pre}}$, MAIT predicts a sample as a member if its loss is smaller than $\tau_{\mathrm{pre}}$.
The reported MIA rate is then the fraction of target samples classified as members:
\[
\mathrm{MIA}(\boldsymbol{\theta};\tau_{\mathrm{pre}})
=
\frac{1}{|\mathcal{D}_{\mathrm{tgt}}|}
\sum_{(\mathbf{x},y)\in\mathcal{D}_{\mathrm{tgt}}}
\mathbb{I}\!\left[\ell(\mathbf{x},y;\boldsymbol{\theta}) < \tau_{\mathrm{pre}}\right].
\]

A lower $\mathrm{MIA}(\boldsymbol{\theta};\tau_{\mathrm{pre}})$ indicates that the target data becomes harder to distinguish as training members and better unlearning.

\end{document}